\documentclass[11pt,a4paper,reqno, dvipsnames]{article}

\usepackage{PRIMEarxiv}

\usepackage[utf8]{inputenc} 
\usepackage[T1]{fontenc}    
\usepackage{hyperref}       
\usepackage{url}            
\usepackage{booktabs}       
\usepackage{amsfonts}       
\usepackage{nicefrac}       
\usepackage{microtype}      
\usepackage{fancyhdr}       
\usepackage{graphicx}       
\usepackage{mathtools, amsmath, amsthm, amsfonts, amssymb} 
\usepackage{bm}
\usepackage{cleveref}
\crefname{section}{§\hspace{-0.1cm}}{§§}
\Crefname{section}{§}{§§}
\usepackage{caption}
\usepackage{subcaption}
\usepackage{tikz} 
\usetikzlibrary{arrows.meta} 
\usetikzlibrary{positioning, shapes.multipart, arrows}
\usetikzlibrary{shapes, fit}
\usepackage{pgfplots}
\pgfplotsset{compat=1.18}
\usepackage{comment} 

  {\list{}{\leftmargin=#1\rightmargin=#1}\item[]}%
  {\endlist}

\usepackage{xspace}

\usepackage{epigraph}

\usepackage{todonotes}
\usepackage{natbib}
\setcitestyle{aysep={}}
\usepackage[toc,page]{appendix} 
\title{does chatgpt have a mind?
}

\author{
Simon Goldstein \\
  The University of Hong Kong \\
  \texttt{simon.d.goldstein@gmail.com} \\
  \And
  B.A. Levinstein \\
  University of Illinois at Urbana-Champaign \\
  \texttt{benlevin@illinois.edu}\\ 
}

\begin{document}
\maketitle

\begin{abstract}
    This paper examines the question of whether Large Language Models (LLMs) like ChatGPT possess minds, focusing specifically on whether they have a genuine folk psychology encompassing beliefs, desires, and intentions. We approach this question by investigating two key aspects: internal representations and dispositions to act. First, we survey various philosophical theories of representation, including informational, causal, structural, and teleosemantic accounts, arguing that LLMs satisfy key conditions proposed by each. We draw on recent interpretability research in machine learning to support these claims. Second, we explore whether LLMs exhibit robust dispositions to perform actions, a necessary component of folk psychology. We consider two prominent philosophical traditions, interpretationism and representationalism, to assess LLM action dispositions. While we find evidence suggesting LLMs may satisfy some criteria for having a mind, particularly in game-theoretic environments, we conclude that the data remains inconclusive. Additionally, we reply to several skeptical challenges to LLM folk psychology, including issues of sensory grounding, the ``stochastic parrots'' argument, and concerns about memorization. Our paper has three main upshots. First, LLMs do have robust internal representations. Second, there is an open question to answer about whether LLMs have robust action dispositions. Third, existing skeptical challenges to LLM representation do not survive philosophical scrutiny.
\end{abstract}

\section{Introduction}

Recent developments in AI are stunning. Large language models like ChatGPT can generate text that is fluent, accurate, and responsive to human questions. These advances have sparked a fundamental question: Do these AI systems possess minds?

To address this broad question, we focus on a more specific aspect of the mental: folk psychology. The key question we explore is whether LLMs have beliefs, desires, and intentions. In other words, do LLMs have goals about what to do, a perspective on what the world is like, and plans for achieving their goals given what the world is like? 

Why care about LLM folk psychology? There are at least three reasons. First, folk psychology is the primary lens that humans use to understand the behavior of agents. When we interact with one another, we consistently try to explain what is happening in terms of beliefs and desires (\cite{sep-folkpsych-theory}). As LLMs become increasingly capable, we will increasingly interact with them. It is worth figuring out whether we can genuinely use beliefs and desires to understand these interactions, or whether instead beliefs and desires would be at best an elaborate metaphor. Second, folk psychology is relevant for moral patiency. When we ask what it takes to be the kind of entity that can be harmed, one important answer appeals to the satisfaction of desires (\cite{Heathwood2016-HEADT}). More may be required too, such as full-fledged consciousness; but folk psychology will play a role. Third, folk psychology is relevant to AI safety. As AI systems become more powerful, many have worried that they may systematically pursue goals that conflict with humanity (\cite{russell2019human}). But much of this discussion implicitly assumes that AI systems will have goals and a perspective about how to achieve those goals. If there are important barriers for AI systems to possess a folk psychology, this may complicate our understanding of what it would take for AI systems to be safe.

Our approach is simple. The question of folk psychology has two key aspects: internal representations and dispositions to act. We'll explore in detail whether LLMs possess each aspect of a folk psychology.

The first key question is whether LLMs have robust internal representations of the world. Here, our strategy in \cref{sec:Theories_of_rep} will be to survey various philosophical theories of representation, and see whether LLMs satisfy the theories. We'll look at a range of conditions on representation, including: (i) that the system has internal states that \emph{carry information} about the world; (ii) that these internal states are \emph{causally effective} at producing the system's behavior; (iii) that  these internal representation satisfy \emph{folk patterns of reasoning}; (iv) that the \emph{structure} of these internal representations mirrors the structure of what they represent; and (v) that the information-carrying capacity of these representations emerged from some kind of \emph{selective, evolutionary process}. In each case, we'll draw on recent research in machine learning about LLM \emph{interpretability}, which suggests that these systems satisfy the relevant condition on mental representation.


But internal representations alone are not sufficient for a full-blown folk psychology. In order to possess beliefs and desires, the second key question is whether LLMs have robust dispositions to perform actions. If LLMs have both internal representations and action dispositions, then they will have a folk psychology. In particular, nearly every theory of belief and desire will explain these mental states in terms of some combination of internal representations and dispositions to act. In \cref{sec:folk_psych}, we explore two of the most prominent traditions of theorizing about belief and desire, interpretationism and representationalism. In each case, we suss out what kinds of action dispositions the theory requires of LLMs. Then we critically assess whether LLMs satisfy the relevant condition. The crucial question for LLMs will be whether their linguistic outputs are \emph{stable} enough to be best explained as promoting goals. Our conclusion in \cref{sec:folk_psych} will be tentative. We argue that the behavior of LLMs in game environments is suggestive of the kinds of rich plans of action required for belief/desire psychology. But the data is not decisive. 

Our third goal of the paper, in \cref{sec:skepticism}, will be to refute some of the existing skeptical challenges to LLM folk psychology. Here, we'll engage directly with three challenges. The first challenge, \emph{symbol grounding}, raises the following question: if the only inputs to a language model are strings of text, rather than rich perceptual experiences or feedback from motor actions, how can language models understand prompts about the external world? The second challenge is that LLMs do not represent the world because they are not trained to do so; instead, they are merely \emph{stochastic parrots}, trained to predict the next word. The third challenge is that LLMs do not represent the world because their behavior can be fully explained by an alternative theory, according to which they rely on \emph{memorization} and shallow shortcuts. In each case, we'll argue that the challenge does not survive philosophical scrutiny.

Overall, then, our conclusions for future research run as follows. First, we think there are interesting questions about exactly how LLMs represent the world, and what in the world they represent. But we think overall there is quite strong evidence that they do so. Second, there is a rich debate to be had about whether LLMs genuinely have beliefs and desires about the world, in the sense that is involved in assembling complex plans of action. Third, we suggest that many of the existing skeptical challenges to LLMs deserve more careful development, as they possess major shortcomings.  Before we turn to our main claims, we'll end this section by briefly summarizing how LLMs work.

\subsection{Large Language Models}

At the heart of modern LLMs lies the transformer architecture (\cite{vaswani2023attention}). Although many variants exist, we'll here focus on the mechanics of decoder-only models such as GPT 3.5.\footnote{More advanced models have many architectural changes, but the exact details are not public.} 

When you feed a prompt to a model, it makes a prediction about what comes next. For example, if you feed it ``The cat sat on the,'' it will assign a probability to each possible next token.\footnote{%
    Tokens can be words, subwords, numerals, punctuation, etc. For our purposes, we can just think of tokens as words. 
} %
It will assign some probability to ``aardvark,'' some probability to ``banana,'' some probability to ``mat,'' and so on. 

To compute these probabilities, each token is converted into an initial \textit{embedding}—a vector, or long list of numbers, which encode ``The,'' ``cat,'' ``sat,'' ``on,'' and ``the.'' The initial embedding carries information about the corresponding token, but it doesn't at first carry any information about surrounding tokens. For example, the initial embedding for ``cat'' does not encode the fact that ``The'' precedes it. 

The initial embeddings for each token are then transformed and updated across a large number of layers. At each layer, the embedding for a given token is first updated through the mechanism of self-attention. Self-attention allows the embedding for a given token to ``attend to'' itself and earlier tokens in the sequence. For example, the token for ``sat'' might attend to ``cat'' at a given layer. The embedding for ``sat'' could then be updated to represent the fact that ``cat'' was the immediately preceding token or, perhaps, to encode somehow that ``cat'' was the subject. In other words, after paying attention to the information of earlier tokens and itself, the embedding for ``cat'' is updated to include contextual information about the surrounding tokens in the prompt. We call this new embedding a \textit{contextual embedding}. These new embeddings are then refined further using something like a multi-layer perceptrons (MLPs) before being fed forward into a new layer. 

\begin{figure}
\centering
\begin{tikzpicture}[
  every node/.style={font=\rmfamily, align=center, inner sep=0.2em, scale=0.8},
  token/.style={rectangle, draw, minimum width=1.2cm, minimum height=0.6cm},
  embedding/.style={rectangle, draw, fill=blue!20, minimum width=1.2cm, minimum height=0.8cm},
  attention/.style={rectangle, draw, fill=red!20, minimum width=1.2cm, minimum height=0.8cm},
  MLP/.style={rectangle, draw, fill=green!20, minimum width=1.2cm, minimum height=0.8cm},
  output/.style={rectangle, draw, minimum width=3cm, minimum height=0.8cm},
  arrow/.style={->, >=stealth, line width=0.5pt}
]
\foreach \word [count=\i] in {The, cat, is, on, the} {
  \node[token] (input-\i) at (\i*1.5, 0) {\word};
}
\foreach \i in {1,...,5} {
  \node[embedding] (emb-\i) at (\i*1.5, -1.5) {$e_\i$ \\ {$\langle \textcolor{gray}{\tiny\bullet}, \textcolor{gray}{\tiny\bullet}, \textcolor{gray}{\tiny\bullet} \rangle$}};
  \draw[arrow] (input-\i) -- (emb-\i);
}
\foreach \i in {1,...,5} {
  \node[attention] (att1-\i) at (\i*1.5, -3) {Att};
  \node[MLP] (MLP1-\i) at (\i*1.5, -4.5) {MLP};
  \draw[arrow] (emb-\i) -- (att1-\i);
  \draw[arrow] (att1-\i) -- (MLP1-\i);
}
\foreach \i in {1,...,5} {
  \foreach \j in {1,...,5} {
    \ifnum\i=\j\else
      \draw[red, dashed, opacity=0.3] (att1-\i) to[bend left=10] (emb-\j);
    \fi
  }
}
\foreach \i in {1,...,5} {
  \node[embedding] (emb2-\i) at (\i*1.5, -6) {$e'_\i$ \\ {$\langle \textcolor{gray}{\tiny\bullet}, \textcolor{gray}{\tiny\bullet}, \textcolor{gray}{\tiny\bullet} \rangle$}};
  \draw[arrow] (MLP1-\i) -- (emb2-\i);
}
\foreach \i in {1,...,5} {
  \node[attention] (att2-\i) at (\i*1.5, -7.5) {Att};
  \node[MLP] (MLP2-\i) at (\i*1.5, -9) {MLP};
  \draw[arrow] (emb2-\i) -- (att2-\i);
  \draw[arrow] (att2-\i) -- (MLP2-\i);
}
\foreach \i in {1,...,5} {
  \foreach \j in {1,...,5} {
    \ifnum\i=\j\else
      \draw[red, dashed, opacity=0.3] (att2-\i) to[bend left=10] (emb2-\j);
    \fi
  }
}
\node[output] (output) at (4.5, -10.5) {Next Token Probabilities};
\foreach \i in {1,...,5} {
  \draw[arrow] (MLP2-\i) -- (output);
}
\node[left] at (-0.5, 0) {Input};
\node[left] at (-0.5, -1.5) {Initial\\Embeddings};
\node[left] at (-0.5, -3.75) {Layer 1};
\node[left] at (-0.5, -6) {Contextual\\Embeddings};
\node[left] at (-0.5, -8.25) {Layer 2};
\node[left] at (-0.5, -10.5) {Output};
\end{tikzpicture}
\caption{Simplified two-layer transformer architecture processing ``The cat is on the''. Each word is initially converted to an embedding vector. In each layer, self-attention (Att) allows words to attend to each other, followed by a multi-layer perceptron (MLP). After the first layer, new contextual embeddings are created. The final layer produces probabilities for the next token.}
\label{fig:transformer_architecture}
\end{figure}
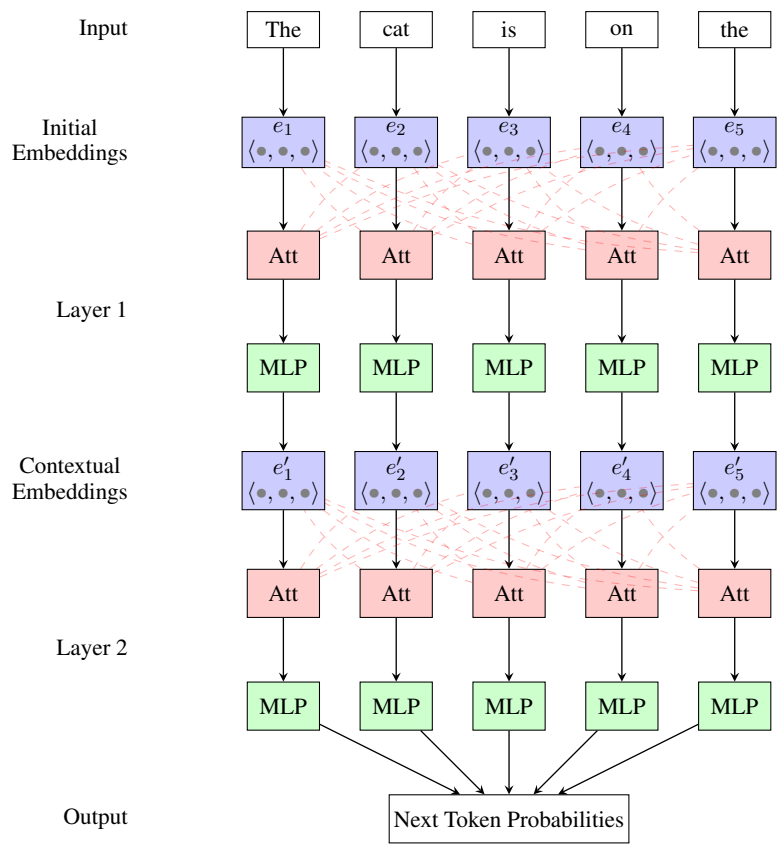

After the embeddings are passed through all the layers of the model, the model uses the final contextual embedding for a token to predict what the next token will be. To generate more and more text, we can then select some token assigned relatively high probability, tack it onto the initial prompt, and then feed the new augmented sequence to the model again. For an illustration, see \cref{fig:transformer_architecture}.

Transformer models get extremely good at generating plausible text via training. Training for retail models like ChatGPT comes in two main phases. In the first phase, we take lots of pre-existing text, feed an initial segment of it to the model, and then have the model generate probabilities about what comes next. We then tweak the parameters of the model via gradient descent to make more accurate predictions. For instance, if the model is fed ``Happy families are all alike; every unhappy family is unhappy in its own'' it will learn to generate a higher probability for ``way'' as the next token. The model's parameters are adjusted through gradient descent to improve its predictions over time.

Eventually, the model gets very good at predicting what comes next in text. But it is still not especially conversational or useful. So, the second (more optional) phase of training, called ``fine-tuning,'' involves tweaking the model further to be a better conversational agent. We will omit the details of this portion of training here, but the essential idea is to get humans or other AI to rate various responses for quality and then push the model to be more disposed to generate high quality responses.\footnote{
    The most common method is reinforcement learning from human feedback (RLHF) \citep{christiano2023deepreinforcementlearninghuman}, but there are a number of alternatives such as reinforcement learning from AI feedback and direct preference optimization.
} 

Note, at this point, some initial obvious reasons for skepticism about LLMs. LLMs are trained, essentially, to make good predictions about sequences of text and then to tell people what they want to hear. There is no direct pressure to represent the world nor to represent truth. Instead, the immediate pressure on LLMs is to find plausible or pleasant continuations of prompts. Furthermore, pure LLMs only connect with the world via textual embeddings. They have some initial embedding for words like ``rainbow'' or ``hammer,'' and they manipulate these embeddings based on context, but they've never actually seen a rainbow nor manipulated a hammer.

There is, then, an important question of whether LLMs can or do end up representing the world as a means toward their training objective.  While they manipulate embeddings based on context, their lack of direct sensory experience and the focus on plausibility rather than accuracy raise reasons for skepticism about their representational capabilities.

\section{Theories of representation}
\label{sec:Theories_of_rep}

Can LLMs represent the world? In this section, we explore this question by examining various philosophical theories of representation. We aim to demonstrate that LLMs meet many of the criteria set by these theories. In particular, we'll highlight a series of recent studies in AI interpretability research and related behavioral research that connect closely to various philosophical theories of representation.

We focus primarily on naturalistic theories of representation. These theories explain representation in terms of physical processes. They differ from other theories, such as those proposing that representation is primitive (\cite{Boghossian1990-BOGTSO}) or dependent on primitive phenomenal properties (\cite{Graham2007-GRACAI}). While non-naturalist theories may allow for  LLM representation, it is difficult to say whether AIs could have primitive representational or phenomenal properties. 

Our focus is on whether LLMs have internal states that represent the world by having truth conditions. A representation can truly depict the world if the world is a certain way and falsely if it is another. We will also consider if LLMs have internal states that refer to objects or properties in the world.

Importantly, our main interest is not the text produced by LLMs but whether LLMs have mental states that represent the world. Our hypothesis is that the \textit{activations} of LLMs---the patterns of internal neural activity within the network as it processes information---refer to objects in the world and have truth conditions. These activations, which can be thought of as the temporary, computation-specific states of the network, are distinct from the more permanent weights that encode the model's learned knowledge. This question of representation is an important first step in determining whether LLMs have a robust folk psychology with beliefs and desires.

With these questions in mind, we will survey leading theories of mental representation to see if LLMs meet their criteria. We draw on existing surveys, including \cite{sep-content-causal} and \cite{Schulte2023-SCHMCD}. Our goal is to show that, according to these leading theories, there is strong evidence that LLMs indeed represent the world.

We will consider five key conditions on mental representations, and argue that LLMs satisfy each one:

\begin{itemize}
    \item \textbf{Information carrying}: Informational theories posit that mental representation requires internal states to \emph{carry information} about the external world, typically through probabilistic connections. We demonstrate how recent advances in AI interpretability, particularly in \emph{probing} techniques, provide compelling evidence that LLM internal embeddings carry such world-relevant information.
   
   \item \textbf{Causal efficacy}: Fodorian theories demand that representational states be \emph{causally effective} in generating system behavior. We present evidence from recent interpretability studies showing that LLM outputs indeed depend counterfactually on their internal embeddings, satisfying this causal requirement.

    \item \textbf{Folk-psychological reasoning}: Another Fodorian condition requires that reasoning with internal representations follows patterns familiar from folk psychology. We argue that emerging research on \emph{world models} in LLMs reveals their capacity for effective reasoning about the world.

    \item \textbf{Structural isomorphism}: Structural theories of representation require that the internal architecture of representations mirrors the structure of what they represent. We illustrate how recent studies on LLM concepts of color and direction demonstrate that their representations exhibit the requisite structural properties.

    \item \textbf{Selection}: Teleosemantic theories stipulate that genuine representations must emerge from a selective, evolution-like process. We contend that the training methodologies employed in developing LLMs fulfill this selectional criterion.
\end{itemize}

\subsection{Information}

A long tradition of work on representation has appealed to the concept of carrying information. Smoke carries information about fire, mumps carry information about measles, and thermometers carry information about temperature. \cite{Dretske1981-DREKAT} and others have argued that a system can only represent the world if that system carries information about it. 

Philosophers have disagreed about how exactly to define carrying information, but in general different analyses all appeal to probabilistic concepts. According to Dretske, a state carries the information that p if and only if the probability of p given the state is 1, provided that various background conditions obtain. Some theorists have instead focused on states raising the probability of p, rather than making it certain (\cite{Usher2001-USHASR}). Other theorists have focused on more general conditions involving entropy.\footnote{%
        The exact connection between carrying information and representing the world is also debated. For Dretske, a mental state represents that p iff the state carries the information that p during the end of the subject's learning period associated with the state. This doesn't require that whenever a state represents that p, it carries the information that p; for Dretske, carrying information is factive, and so this would rule out misrepresentation. But other notions of carrying information might not be factive, and so could allow a more direct connection between representation and information.
    } %

How can we tell whether LLMs carry information about the world? \cite{HardingForthcoming-HARORI-3} argues  there is a close connection between informational approaches to representation and recent work on probing in AI interpretability research (\cite{alain2018understanding}). In probing, researchers train a separate classifier to take activations as input and make a prediction. The probe takes in some  activations and predicts features of the input. For example, in a visual AI system, a probe might predict whether the system is looking at a cat based solely on  activations, without access to the original input.\footnote{Probes typically use linear classifiers or shallow neural networks trained on model activations to predict specific features. For details, see \citep{alain2018understanding}.}

\begin{figure}
\centering
\begin{tikzpicture}[
    node distance=1.3cm,
    box/.style={rectangle, draw, rounded corners, minimum width=4cm, minimum height=1cm},
    vector/.style={rectangle, draw, minimum width=4cm, minimum height=0.8cm},
    arrow/.style={->, >=stealth, thick}
]

\node[box, fill=blue!10] (input) {Input: ``F5 D6 C3 D3 C4 F4 E3''};

\node[box, below=of input, fill=green!10] (llm) {Othello-GPT};

\node[vector, below=of llm, fill=yellow!10] (activation) {[0.3, -0.1, 0.7, ...]};

\node[box, below=of activation, fill=red!10] (probe) {Probing Classifier};

\node[box, below=of probe, fill=blue!10] (output) {
    \begin{tabular}{c}
        Output: Square D4 is blank 
    \end{tabular}
};

\draw[arrow] (input) -- (llm);
\draw[arrow] (llm) -- node[right] {Activation} (activation);
\draw[arrow] (activation) -- (probe);
\draw[arrow] (probe) -- (output);

\node[left=0.2cm of llm] {LLM Processing};
\node[left=0.2cm of probe] {Feature Extraction};

\end{tikzpicture}
\caption{Illustration of the probing process using Othello-GPT. The LLM processes the input sequence of Othello moves, generating activations. A separate probing classifier is trained to predict specific features (e.g., the state of a particular square) from these activations.}
\label{fig:probing_illustration}
\end{figure}
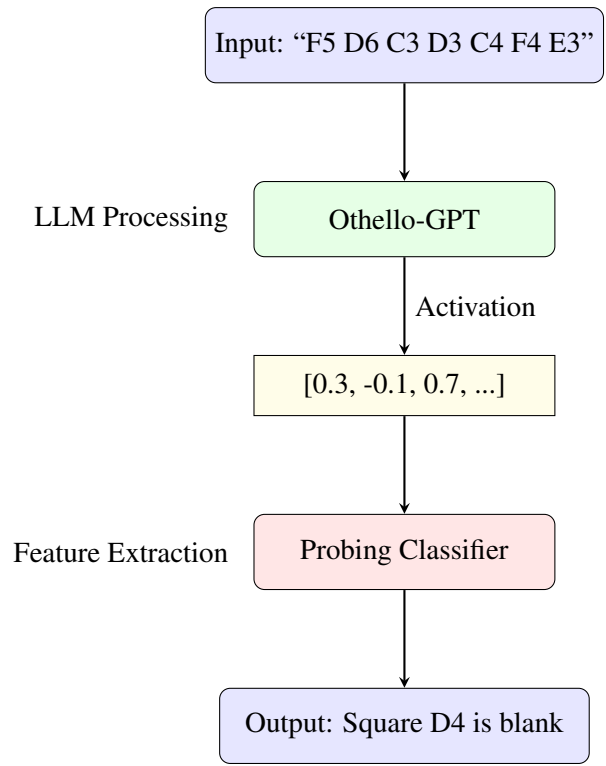

A compelling example of using probes to discover LLM representation comes from \cite{li2022emergent}. We'll use this example as a case study below. Li et al trained an LLM on sequences of moves in the 8x8 board game Othello, using only lists of moves (like F5 D6 C3 D3 C4 F4 E3) without describing the rules or the board.  Othello is a simpler game than chess, but there are far too many possible moves in general for an LLM simply to memorize all legal game states. Despite this, the trained LLM, dubbed Othello-GPT, tended to output legal moves with high probability. To understand how, the authors used probes to identify possible internal representations of the board. By comparing the model's internal states to the actual board, they trained probes to guess whether each of the sixty-four squares was black, white, or blank. The probes achieved remarkable accuracy with an error rate of only 1.7\%. Similar probing techniques have been used to understand how models represent grammatical case, number, tense, and more (e.g., \cite{giulianelli2021hood}). \Cref{fig:probing_illustration} illustrates the probing process.

Informational theories of representation  make sense of the relevance of probes for mental representation. If the classifier can correctly determine whether a cat is present almost 100\% of the time, and the classifier is only using the  activations of the system to make its prediction, then those  activations likely represent the cat. At the very least, they carry mutual information with the presence of cats. If mental representation is a matter of carrying information, then there is no special barrier to LLMs representing the world. 

\subsection{Causal powers}


\cite{Fodor1975-FODTLO} imposed a series of conditions on genuine representations, one of which is that they must have genuine causal powers. The key question here is whether LLM outputs are robustly caused by the activations discovered by probes or whether these apparent representations are actually epiphenomenal.

To make this concern concrete, return to Othello-GPT. Suppose Othello-GPT itself does not represent the state of the board at all. However, a clever probe could read off from its activations alone what the history of moves was. For instance, if F5 D6 C3 D3 C4 F4 E3 were fed to Othello-GPT, the probe could learn just from its internal state that F5 D6 C3 D3 C4 F4 E3 was the input. If the probe also learned (via its own training) what the rules of Othello were, it could determine whether each square was black, white, or blank, even if Othello-GPT itself didn't represent that fact.

Alternatively, we can imagine that Othello-GPT does represent the board state, but the probe found some other way to determine whether its square was black, white, or blank using information from Othello-GPT's activations that Othello-GPT itself does not use.
To investigate this question, interpretability researchers use methods of causally intervening on a model's activations to test whether supposed representations are actually relevant or useful to the model.\footnote{%
    There are many different methods of intervention to understand and manipulate neural network representations. The simplest is ablation, where entire neurons are deactivated to assess their importance in the model's performance. More sophisticated approaches include:
    \begin{itemize}
        \item Iterated Nullspace Projection (\cite{ravfogel2020null}): This method iteratively projects embeddings into a nullspace to remove the influence of specific concepts, effectively isolating and erasing targeted information.
        \item Least Squares Concept Erasure (LEACE) (\cite{belrose2023leace}): LEACE identifies and removes concept-specific information from embeddings using a least squares optimization approach, ensuring that the targeted concept is no longer represented in the neural activations.
        \item Causal Tracing (\cite{meng2023locating}): This method tracks the flow of information through a model to determine the causal impact of specific components or representations. By intervening in the causal pathways, researchers can assess the importance and role of particular features in the model's decision-making process.
    \end{itemize}
    These techniques allow researchers to surgically target and manipulate information encoded in embeddings, providing deeper insights into the functioning and interpretability of neural networks.
    } %
For instance, if we find a successful probe for a square in Othello-GPT, we can determine what elements of Othello-GPT's activations the probe is using to identify whether a square is black, white, or blank. We can then hand-edit the activation to see what happens.

\begin{figure}
    \centering
    \begin{subfigure}[b]{\textwidth}
        \centering
        \includegraphics[width=0.9\textwidth]{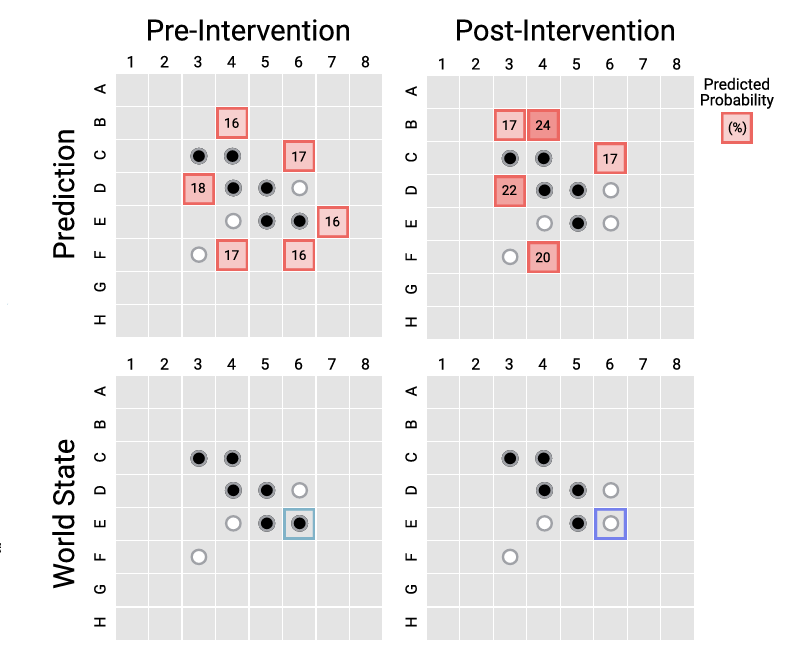}
        \caption{Othello board state and model predictions before and after intervention}
        \label{fig:othello_board}
    \end{subfigure}
    \vspace{1em}
    \begin{subfigure}[b]{\textwidth}
        \centering
        \includegraphics[width=0.9\textwidth]{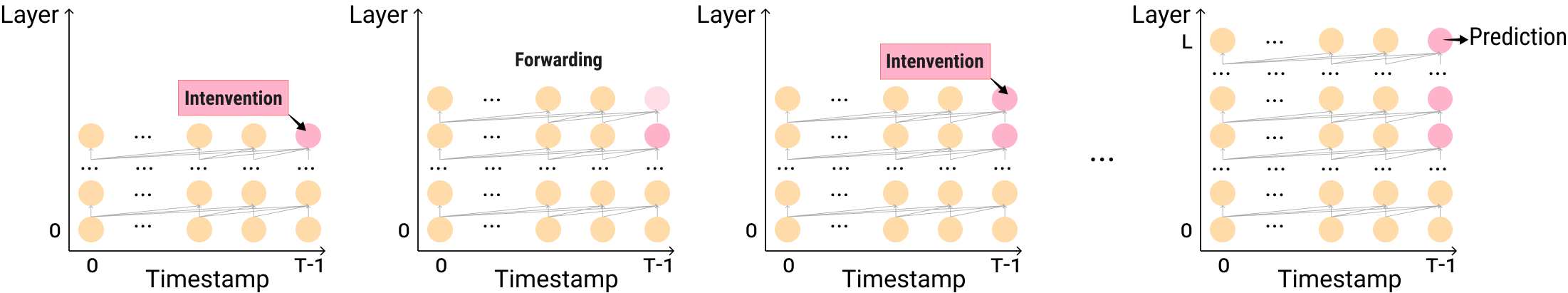}
        \caption{Activation patching process across model layers}
        \label{fig:activation_patching}
    \end{subfigure}
    \caption{Activation patching in Othello-GPT. (a) The top panel shows the Othello board state and model predictions before and after intervention. The upper row in each state displays the model's move predictions with associated probabilities, while the lower row shows the actual board state. Pre-intervention, the model correctly predicts legal moves. Post-intervention, the model's predictions change. (b) The bottom panel illustrates the process of activation patching across different layers and timestamps of the model. The intervention at a specific layer and timestamp propagates through subsequent layers, ultimately affecting the final prediction. This demonstration shows how altering internal representations can causally influence the model's outputs, even leading to illegal move predictions in the context of the game state shown above. Both depictions are adapted from \cite{li2022emergent}.}
    \label{fig:othello_activation_patching}
\end{figure}

Suppose, in an oversimplified example, that the probe notices that when the first coordinate of an embedding vector for a square is 1, the square is black; when it is 0, it is blank; and when it is $-$1, it is white. We can feed the model a prompt making the square black, setting the first coordinate of the relevant embedding vector to 1. Then, we can manually change this coordinate to 0 or $-$1. If the model's output changes as expected, we have strong evidence that the representation we found is one the model itself uses to track the board state. \Cref{fig:othello_activation_patching} illustrates the process of causal intervention in the Othello-GPT model. This demonstrates how altering specific internal representations can change the model's outputs, providing evidence for the causal efficacy of these representations.

As another example, consider finding a potential representation of grammatical number in an LLM. Suppose we feed the LLM the prompt `I ate these'. Without hand-editing, `apples' should receive a higher probability than `apple' as the next token. But if we hand-edit the representation of `these' from plural to singular, we can see if `apple' becomes more likely than `apples'. If so, we have causal evidence that our supposed representation is used by the model.

In the case of Othello-GPT, causal interventions were effective showing that the probes did in fact find the model's representation of the board's state. In other cases, similar causal methods have been used to find representations of grammatical number, subjecthood, and factual associations (\cite{giulianelli2021hood}, \cite{meng2023locating}). We thus have direct evidence of the causal effectiveness of LLM representations. 

\subsection{Folk patterns of reasoning}

So far, we've argued that LLM activations carry information and causally influence LLM outputs. But this alone may not be enough for genuine representation. For example, \cite{Fodor1975-FODTLO} argued that genuine mental representations have to have special kinds of causal powers: the representations have to influence the system's behavior in ways that match the laws of folk psychology. The representations have to make sense of the world, causing outputs in a way that resembles ordinary reasoning. Given our interest in whether LLMs have folk psychological states, this theory of representation is especially important for our purposes.

One way interpretability researchers have investigated this kind of condition is through the idea of world models. A world model, in the context of AI, refers to an internal representation of how the external world operates, including objects, their properties, and the causal relationships between them. World models require coherent and consistent representations across contexts and a degree of abstraction that allows LLMs to generalize from particular cases to more general situations.  For LLMs, these models are constructed purely from textual data, raising important questions about their nature and limitations.

The importance of world models lies in their potential to bridge the gap between mere pattern recognition and genuine understanding. If LLMs possess robust world models, it suggests they have developed representations that go beyond simple statistical associations, potentially supporting attributions of beliefs and reasoning capabilities.

\cite{li2022emergent} suggest their experiments with Othello-GPT are strong evidence of LLMs' ability to create world models, since Othello-GPT has some coherent internal model of the board state that allows it to predict the next move.  Indeed, Othello-GPT can easily generalize to make predictions about new boards its never seen before and can even model permissible moves in impossible boards---i.e., boards with states that can never be reached through a legal initial series of moves. Recent work by \cite{vafa2024evaluating} builds upon and extends this approach, proposing new evaluation metrics for world model recovery inspired by the Myhill-Nerode theorem from language theory. Their study encompasses not only game environments like Othello but also navigation tasks and logic puzzles that can be captured by a finite state automaton. Their research demonstrated that while language models can perform well on existing diagnostics, their underlying internal models vary in levels of coherence.

However, the rules of Othello and those studied by \cite{vafa2024evaluating} are essentially a kind of language that renders some ``moves'' grammatical and others not. It's not clear that modeling Othello is sufficient evidence of LLMs' having causal abstractions of physical processes that are robust enough to count as world models. 

\cite{musker2023testing} explore whether large language models build causal models in order to understand the meaning of words. They rely on the HIPE theory, developed to understand human lexical concepts, in connection with artifact terms like ``mop'' and ``pencil'' (\cite{chaigneau2004assessing}). According to this theory, when human language users decide whether an artifact term like ``mop'' can correctly apply to an object, they rely on an implicit causal model: 

\begin{quote} The object's design history and the user's goal are distal causes in the CM, while the object's physical structure and the user's actions with respect to it are proximal causes in the CM. Thus, HIPE predicts that, for example, both the physical structure of an object (e.g., having a handle and something absorbent on one end) as well as the reason the object was originally created (e.g., for wiping up water) should affect how appropriate it is to call the object a `mop', but that the latter should have a minimal effect when the former is fully specified (Musker and Pavlick 2023).
\end{quote}

The idea is that the status of later nodes `screen off' earlier nodes when judging whether something is an artifact. To test this theory, \cite{chaigneau2004assessing}  built vignettes that manipulated various nodes in the causal model, and explored which of the nodes influenced language users' judgments about whether an object counted as a type of artifact. For example, one such vignette was: 

\begin{quote} One day Jane wanted to wipe up a water spill on the kitchen floor, but she didn't have anything to do it with. So she decided to make something. [...] The object consisted of a bundle of thick cloth attached to a 4-foot long stick. Later that day, John was looking for something to wipe up a water spill on the kitchen floor. [...] He grabbed the object with the bundle of thick cloth pointing downward and pressed it against the water spill (\cite{musker2023testing}). 
\end{quote} 

\cite{musker2023testing} applied this theory of lexical concepts to large language models. Musker and Pavlick found that GPT-4 behaved very similarly to human subjects when evaluating counterfactual vignettes. Compromising any factor in the model led to a negative effect on GPT-4 deciding that something counted as an artifact. Compromising proximal features had a larger effect on GPT-4's judgments than compromising distal features.\footnote{For example, in one vignette (labeled ``pencil object, compromised action scenario'') an object is designed to be used as a pencil (satisfying the goal condition), but the user fails to successfully use the object to write on a piece of paper (violating the action condition).} This potentially suggests that GPT-4 built causal models in order to deploy artifactual lexical concepts, in a way structurally similar to humans.\footnote{%
    For more on the emergence of causal models in LLMs, see \cite{forbes2019neural}, \cite{da2019cracking}, \cite{ettinger2020bert}, \cite{petroni2019language}, and \cite{kassner2020negated}. Musker and Pavlick themselves shy away from interpreting these results as showing that GPT-4 genuinely builds causal models of artifacts (p. 7). Instead, they suggest that more work is needed to identify methods for probing inner models. After all, their methodology relies solely on GPT-4's responses to text vignettes. But we have already seen that other work in interpretability research has used probing and other paradigms to uncover world models in the internal representations of LLMs. In this setting, even the behavioral evidence from Musker and Pavlick can potentially be interpreted as revealing inner causal models associated with LLM lexical concepts. See \cite{yildirim2024task} for further work exploring the philosophical upshots of world models in LLMs.}

\begin{figure}
    \centering
    \begin{subfigure}[b]{\textwidth}
        \centering
        \includegraphics[width=0.9\textwidth]{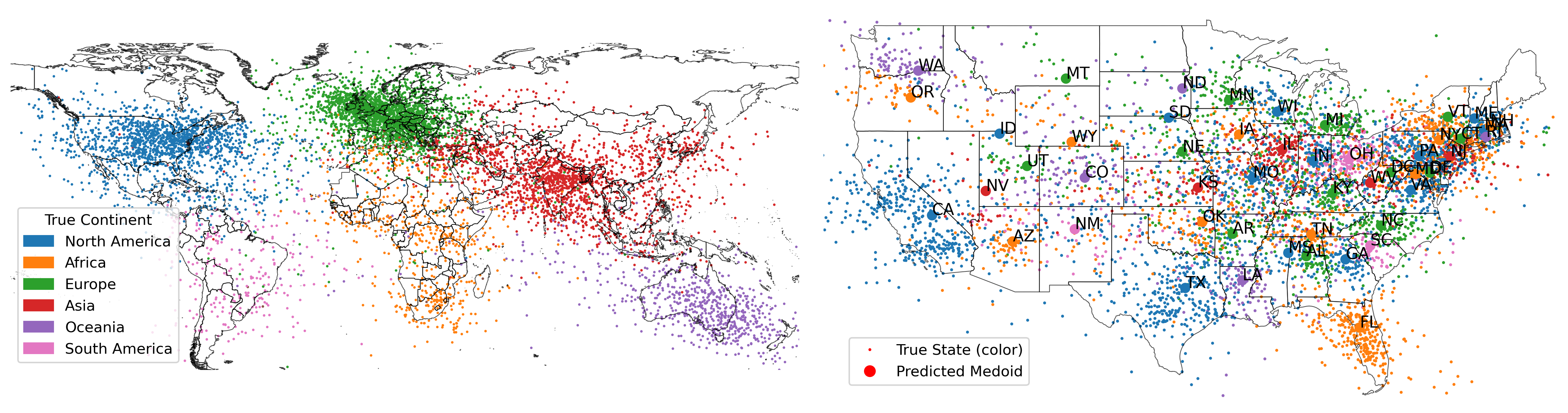}
        \label{fig:Tegmark_world_us}
    \end{subfigure}
    \vspace{1em}
    \begin{subfigure}[b]{\textwidth}
        \centering
        \includegraphics[width=0.9\textwidth]{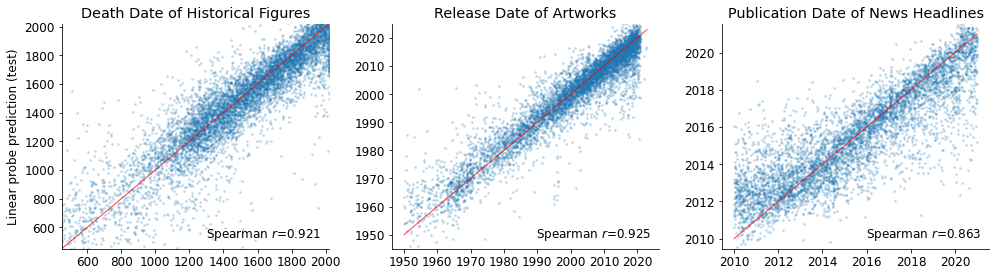}
        \label{fig:Tegmark_time}
    \end{subfigure}
    \caption{Spatial and temporal representations in Llama-2-70b. Each point corresponds to the layer 50 activations of the last token of a place (top) or event (bottom) projected onto a learned linear probe direction. The clear structure in these projections, closely matching real-world geography and chronology, demonstrates that the model has learned coherent representations of space and time. All points depicted are from the test set. (Adapted from \citep{gurnee2024language}.)}
    \label{fig:tegmark}
\end{figure}

Recent work by \cite{gurnee2024language}  provides compelling evidence that large language models (LLMs) develop coherent representations of space and time, even when trained solely on text data. By probing the internal activations of Llama-2 models, they discovered that LLMs learn linear representations of spatial and temporal information across multiple scales. These representations are unified across different entity types (e.g., cities and landmarks) and robust to variations in prompting. Remarkably, they identified individual ``space neurons'' and ``time neurons'' that reliably encode spatial and temporal coordinates. \Cref{fig:tegmark} provides striking visual evidence of how LLMs develop structured representations of space and time. The clear organization of locations and events in the model's internal space suggests that these representations go beyond mere statistical associations.

Another skeptical challenge to LLM reasoning concerns poor performance in LLMs like GPT 3.5 when such models are asked to perform arithmetic operations.  In order to genuinely represent mathematical information on a Fodorian view, the LLMs would need a series of representations that transform in law-like ways that reflect folk patterns of reasoning. Instead of genuinely representing mathematical information, the skeptics argue, these models are instead engaged in simplistic statistical pattern matching or simply memorize their training data. 

\cite{nanda2023progress} explored internal computations in a LLM trained to perform modular addition. They found that rather than merely memorizing answers or relying on statistical patterns, the LLM implemented a particular `Fourier multiplication algorithm' for computing the sum: simply put, ``they perform this task by mapping the inputs onto a circle and performing addition on the circle.''\footnote{%
        See \cite{zhong2023clock} for an alternative algorithm some LLMs learned to perform modular arithmetic. 
    } 

During training, the models initially overfitted the data by memorizing specific examples. However, over time, they transitioned to a generalizable understanding of modular addition, demonstrating an internal shift from memorization to systematic reasoning.\footnote{%
    There are many more questions that can be asked about the extent to which LLM representations satisfy various folk reasoning patterns. Recent discussion, for example, has highlighted the `reversal curse', where LLMs are competent with the phrase A is B, while failing to understand the phrase B is A (\cite{berglund2024reversal}). There has been rich recent discussion about these sorts of issues, caught up with the question of whether LLM representations exhibit the kind of full-blown compositional structure familiar from \cite{Fodor1975-FODTLO}. (For example, see \cite{lake2023human} for a recent argument that LLMs do exhibit full compositional structure.)
    } %

\subsection{Structure}

Structuralist theories of mental representation propose that mental states represent the world only if their internal structure mirrors the structure of the external world  (\cite{Opie2004-OPINTA}). This mirroring creates a network of relationships within the mental states that correspond to the relationships between the objects or properties they represent. As an analogy, maps represent geographic areas by containing a series of symbols whose physical relations on the page mirror the physical distances between locations.

Structuralist theories of mental content are particularly relevant to large language models. Interpretability researchers have explored the activations of LLMs to see whether they stand in patterns of relations with one another that are isomorphic to various worldly features. 

\cite{patel2021mapping} tested GPT-3's concepts of direction and color. In the case of cardinal directions, they exposed GPT-3 to a series of gridworlds, consisting of matrices filled with 0s and a single 1. The task was to identify the direction of the 1 symbol in the gridworld: left, right, etc. They prompted GPT-3 with a series of sample answers, and tested whether GPT-3 could complete the pattern.  
Patel and Pavlick found that GPT-3 had a strong grasp of cardinal directions. First, the LLM could smoothly generalize to a range of gridworld environments: it could correctly describe directions in gridworlds with new lengths and widths. Second, the LLM could smoothly generalize across cardinal directions: even when it was only prompted with northerly or easterly directions, it could correctly answer questions about southerly and westerly located 1s. This showed that GPT-3 had an underlying conceptual space of directions that connected the concepts of north to south, and west to east. Once the LLM could hook up one of these concepts to the gridworld, its underlying structural understanding of direction allowed it to complete the grounding task. Finally, the LLM smoothly generalized to rotated gridworlds. Even when the grids did not map onto our ordinary judgments of left and right, the LLM could quickly generalize from examples. This suggested that the LLM had a stable underlying conceptual space for directions.

Patel and Pavlick ran a similar experiment for color. In this case, they tested whether LLMs could few-shot learn how to map folk color terms to three dimensional RGB scales. Again, although GPT-3 is a text-only model, it nonetheless possessed an underlying conceptual space for color. Even when it was only prompted with examples that associated RGB values with red colors, it could still use this information to correctly predict how to associate RGB values with blue colors. This showed that the LLM had an underlying grasp of the structural relation between red and blue colors: once it learned how to connect red colors to RGB values, the further connection to blue colors could be inferred. 

\begin{figure}
    \centering
    \begin{subfigure}[b]{\textwidth}
        \centering
        \includegraphics[width=0.9\textwidth]{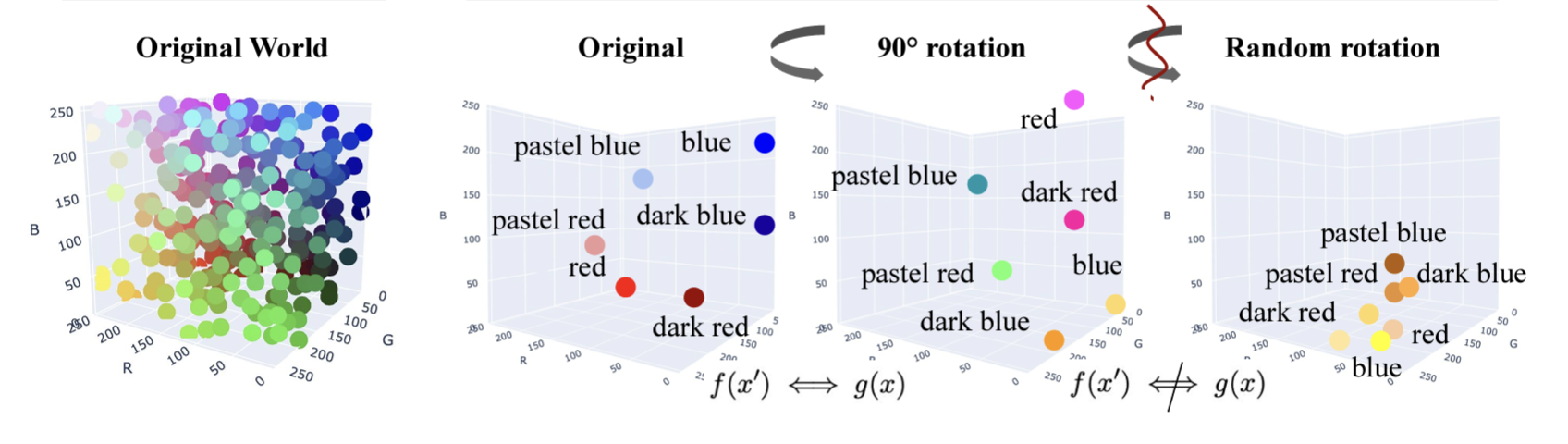}
        \caption{ Isomorphic and Random Rotations in Color Space: The leftmost panel shows the full 3D color spectrum. The three right panels demonstrate how sample colors transform under different conditions: in their original positions, after a 90° rotation (an isomorphic transformation), and after random reassignment. This illustrates how structural relationships between colors are preserved in isomorphic rotations but disrupted in random rotations.}
        \label{fig:color_rotation}
    \end{subfigure}
    \vspace{1em}
    \begin{subfigure}[b]{\textwidth}
        \centering
        \includegraphics[width=0.9\textwidth]{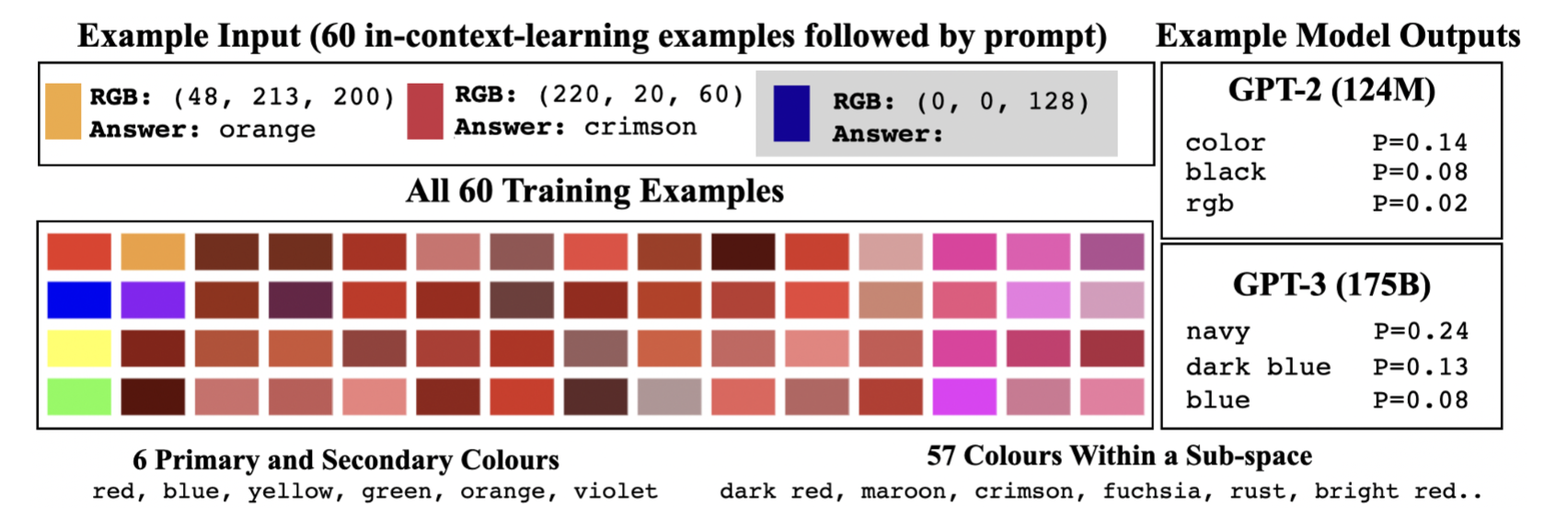}
        \caption{Generalization to Unseen Color Terms: The left panel shows example training data, including primary, secondary, and shades of red. The right panel demonstrates model outputs for an unseen color (navy blue). Importantly, navy blue is in a separate color subspace from any of the colors used for training.}
        \label{fig:color_subspace}
    \end{subfigure}
    \caption{These figures together illustrate the study's approach to testing language models' understanding of color terms. While panel (b) suggests the model's ability to generalize to unseen colors, panel (a) shows how rotated color spaces were used to control for potential memorization of RGB-to-name mappings. This combined approach helps distinguish between true generalization and mere memorization of training data. (Adapted from \citep{patel2021mapping}.}
    \label{fig:Pavlick_color}
\end{figure}

In both the case of direction and color, Patel and Pavlick's experiment demonstrated that LLMs develop rich networks of representations, with structural relationships that resemble the analogous human concepts. This result fits related interpretability research from \cite{abdou2021language}, which found that text-only language models create a network of embeddings for color terms. These embeddings were found to have structural relationships that mirrored human judgments of similarity between various colors: they found that language model ``representations of color terms that are derived from text only express a degree of isomorphism to the structure of humans' perceptual color space''. For a visual illustration, see \cref{fig:Pavlick_color}.

\begin{figure}
\centering
\begin{tikzpicture}
\begin{axis}[
    width=12cm,
    height=9cm,
    xlabel={},
    ylabel={},
    xmin=-5, xmax=5,
    ymin=0, ymax=4,
    xtick=\empty,
    ytick=\empty,
    axis lines=middle,
    legend pos=north west,
    legend cell align=left,
]
\addplot[->,blue,ultra thick] coordinates {(0,0) (-1,2)} node[above] {man};
\addplot[->,blue,ultra thick] coordinates {(0,0) (3,3)} node[above] {king};
\addplot[->,red,ultra thick] coordinates {(0,0) (-3,2)} node[above] {woman};
\addplot[->,red,ultra thick] coordinates {(0,0) (1,3)} node[above] {queen};
\addplot[gray, ultra thick, dashed] coordinates{(3,3) (4,1)};
\addplot[gray, ultra thick, dashed] coordinates{(4,1) (1,3)};

\node[rotate=285, gray, anchor=south] at (3.4,2.3) {$-$ man};
\node[rotate=310, gray, anchor=south] at (2.8,1.4) {$+$ woman};

\end{axis}
\end{tikzpicture}
\caption{Word embedding vector operations illustrated: king - man + woman $\approx$ queen}
\label{fig:word_embeddings}
\end{figure}
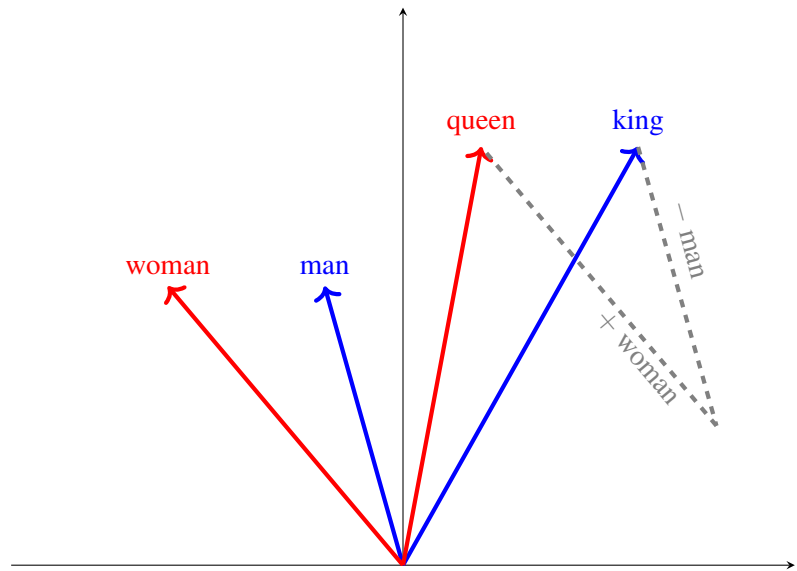

Indeed, the nature of the embedding space allows for a natural representation of many different structural relationships. Word embeddings are dense vector representations of words, where words with similar meanings are located close to each other in the vector space. This spatial arrangement allows LLMs to recognize and generate analogies by identifying patterns and relationships between different word vectors. In the case of the analogy ``man is to woman as king is to queen,'' LLMs can identify this relationship through the arithmetic of word vectors (\cite{vylomova-etal-2016-take}). The vector difference between ``man'' and ``woman'' is similar to the vector difference between ``king'' and ``queen.'' Mathematically, this can be expressed as: king - man $\approx$ queen - woman. For a visual illustration, see \cref{fig:word_embeddings}.

When the LLM searches for a word that satisfies this relationship, it can correctly identify ``queen'' as the word that completes the analogy. Likewise, the model can understand the relationship between a city and its country through a similar mechanism. The vector difference between ``Paris'' and ``France'' is similar to the difference between ``Tokyo'' and ``Japan,'' enabling the model to complete the analogy.

Structural conditions on representational content may not themselves be sufficient to fully explain how mental states have truth conditions.\footnote{%
        See \cite{piantadosi2022meaning}  for an extended discussion of conceptual role semantics in LLMs.
    } %
After all, several different networks of physical properties could stand in the same structural relations, and this would then leave unsettled which of them is the referent of the relevant mental state. But structural conditions can be combined with other conditions to produce a full theory of mental content. For example, one strand of work in philosophy of mind has appealed to structural conditions to explain the particular significance of phenomenal experiences. According to representationalists about phenomenal character, phenomenal experiences supervene on their representational content. According to these theories, there can be no change in how things seem qualitatively without a change in the representational content of one's experiences. Many of these representationalists attempt to explain phenomenal experiences in terms of a special kind of representational content. Here, structural conditions have played an important role. For example, \cite{Rosenthal2005-ROSCAM-8} argued that qualitative experiences are associated with a network of representational contents that are arranged in a similarity space. Consider color concepts. Rosenthal's idea is that color experiences not only represent the world, but also stand in patterns of similarity to one another: red experiences are more similar to orange experiences than to yellow ones. And these similarity patterns are isomorphic to patterns in what color experience represents (for example, wave length). The interpretability research surveyed above suggests that LLMs may satisfy this condition on rich color representation. 

Finally, this work on color concepts is also relevant to our earlier question of whether LLM activations satisfy generalizations from folk psychology. When LLMs successfully make predictions about the relationships between colors or directions they haven't previously encountered, they seem to reason in ways that are similar to the way humans reason. 

\subsection{Teleosemantics}

Another potential necessary condition on representation comes from `teleosemantic' theories of mental content. According to teleosemantic theories, the function of the underlying system generating a mental state determines its content (\cite{Millikan1984-MILLTA}). For example, whether a given mental state counts as a representation of a goldfinch depends on the function of the system that created that mental state. On standard teleosemantic views, the function of a state is determined by evolution: the state has whatever function explains why it was selected for via natural selection (see \cite[p. 35]{Schulte2023-SCHMCD}). 

Teleological conditions can be combined to causal and informational conditions on representation. For example, \cite{Stampe1977-STATAC}'s causal account of content appealed to the idea of causal correlations that obtain in normal conditions. This can be fleshed out in terms of evolution: a perceptual state represents a feature of the world if that state being reliably caused by the feature helps explain why the organism was selected for in natural selection. Similarly, \cite{Neander2013-NEATAI, Neander2017-NEAAMO} argues that perceptual states have content when they have the function of being reliably produced by features of the environment. For example, toads have evolved perceptual states that fire in response to ``small, dark, moving'' objects. These states were selected for by natural selection, because toads with these states were more successful at hunting flies. For this reason, these perceptual states represent small, dark, moving objects. Similarly, informational accounts can be supplemented with the condition that various information-carrying channels were selected for (see \cite{Dretske1981-DREKAT}). 

We argue that teleological constraints on meaning are broadly compatible with AI representation. Although LLMs are not products of biological evolution, they undergo a form of artificial selection during their training process. This selection process optimizes the model's parameters---specifically, its weights---to improve its performance on specific tasks. In teleosemantic theories, the function of a mental state is determined by what it was selected for in the evolutionary process. Analogously, in LLMs, the function of a particular weight configuration is determined by its role in minimizing the loss function during training.

During training, the model's weights evolve in response to the pressure of minimizing loss, much like biological traits evolve under natural selection. This process bears important similarities to natural selection: just as biological traits that enhance fitness are more likely to persist, weight configurations that reduce prediction error are more likely to be retained. The resulting weight structure of the trained model can be seen as encoding the ``functions'' that the model was selected to perform, paralleling how evolved biological structures encode their functions.

The crucial disanalogy to natural selection is that in natural selection mutation is random, while in machine learning weights do not change randomly, but are instead adjusted in the direction of lower loss. But this disanalogy does not seem relevant to teleosemantic theories. In both cases, an underlying optimization process creates a clear sense of a goal, and of normal versus abnormal pursuit of that goal. 

At this point, we've surveyed several different potential requirements on mental representation. We have argued that results about AI interpretability suggest that large language models can meet each of these requirements. We think that when LLMs play games like Othello, for example, they don't merely predict the next word. Instead, they represent the game environment. In particular, as LLMs learn how to complete this task, they build internal models, representations that help them figure out what to do next, for example by representing board states. In this way, we think that interpretability research provides significant reason to reject skepticism about mental representation.

\section{Responding to Skeptical Challenges}
\label{sec:skepticism}

At this point in the paper, we've laid out our positive claims about LLM mental representation. In short, we think that there is a strong case that LLMs possess robust internal representations of the world. 

In this section, we'll respond to three skeptical challenges to LLM folk psychology. Each skeptical challenge targets the claim that LLMs represent the world. (In the next section, we turn to challenges to LLMs possessing robust action dispositions.) In each case, we'll identify potential gaps in the skeptical challenge, and consider how the skeptical challenge is potentially relevant to questions about LLM folk psychology. 

We'll consider three challenges:

\begin{itemize}
\item \textbf{Sensory grounding}: This challenge says that LLM inputs are meaningless because LLMs have no connection to the external world. In response, we'll argue that (i) the challenge relies on overly simplistic causal constraints on mental representation; (ii) the challenge ignores the possibility that LLMs form hypotheses about aspects of the world that are not directly observable to them; and (iii) the challenge is \emph{fragile}, because it relies on properties of LLMs (such as the lack of perceptual inputs) that are not shared by all models. 
\item \textbf{Stochastic parrots}: This challenge says that LLMs do not represent the world because they aren't trained to do so. In response, we'll argue that (i) the challenge overgeneralizes to threaten human cognition; (ii) the challenge ignores that LLMs could represent the external world as a means to predicting the next word; (iii) the challenge ignores the possibility that representation is an \emph{emergent capability} of LLMs; and (iv) the challenge is incompatible with the structure internal representations identified by interpretability research.
\item  \textbf{Memorization}: This challenge claims that LLMs do not represent the world because their behavior can be explained through the alternate theory that they simply memorize text. In response, we'll argue that LLMs possess the ability to generalize robustly from their training data, and then make correct predictions about new questions.
\end{itemize}

We can think of each of these three challenges as targeting one of the conditions on representation we discussed earlier. The sensory grounding challenge claims that representation requires specific kinds of causal connections between the external world and internal LLM states. The memorization challenge claims that representation requires robust models and reasoning abilities, which LLMs supposedly lack. The stochastic parrots challenge claims that representation requires the right kind of functional origin, rooted in training objectives. 

\subsection{Sensory Grounding}

One skeptical challenge concerns sensory grounding. This skeptic says that LLMs don't represent the world because they lack sensory grounding and merely use text: ``a system that is trained only on form [such as an LLM] would fail a suﬃciently sensitive test [for intelligence], because it lacks the ability to connect its utterances to the world'' (\cite{bender-koller-2020-climbing} p. 5188). Pure (text-only) LLMs only see patterns of text. They do not have any outside sensory input. Therefore, according to this challenge, they can't ``break the syntactic circle'' and connect any of the symbols they see to the outside world. As \cite{harnad1990symbol}  defined it, the symbol grounding problem is one of how the semantic interpretation of symbols can be ``intrinsic to the system, rather than just parasitic on the meanings in our heads.'' 

One version of the symbol grounding problem involves causal theories of mental representation. For some theories of mental representation, the right sort of causal connection between the outside world and internal states is required for such states to count as representational in the first place. According to these theories (including \cite{Stampe1977-STATAC} and \cite{Fodor1987-FODPTP}), mental states represent the world when they are causally connected to the world in the right way.\footnote{%
        The precise details of the causal theory differ with each particular adherent. Stampe proposed that a mental state represents the proposition that p iff under optimal conditions, it is causally correlated with p. Stampe's notion of optimal conditions was then spelled out in terms of biologically normal conditions that ``guarantee the well-functioning'' of the mechanisms that produce the mental state. Fodor defended a different, ``asymmetric dependence'' theory. For Fodor, a concept C represents an object O when the concept is lawfully causally correlated with that object, and any other correlations between C and other objects are asymmetrically explained by the connection between C and O. For example, my concept ZEBRA is caused by zebras, but can also be caused by cleverly painted mules. But when ZEBRA is caused by a painted mule, this itself is explained by the underlying causal connection between ZEBRA and zebras. 
    } %
For example, my perceptual experiences of cats tend to be caused by cats, and therefore represent them. In addition, my desire to eat ice cream tends to cause me to eat ice cream, and therefore represents ice cream. Causal theories of representation can explain, for example, why a photograph of an identical twin represents one twin and not the other, despite resembling each twin perfectly.

Naive causal conditions may make trouble for LLM representation. LLMs have activations that are related to cats. But these  activations are not directly caused by cats in the way that our perceptual system directly responds to cats in our environment. Instead, LLMs have learned about cats through training on text, and this text itself has been caused by cats.

There are two ways to address causal skepticism about LLM representation. The first option is to appeal to long causal chains. The second option is to appeal to pluralism about causal structure. Let's consider each in turn.

The first response appeals to long causal chains. In some sense, LLM activations related to cats are not caused directly by cats, in the way that retinal stimulation is directly caused by cats. Instead, LLM activations related to cats are caused by text about cats. But cat text is itself caused by cats. In this way, there is a causal chain from cats to LLM activations about cats. So it isn't clear that even naive causal conditions on representation block LLMs from genuine reference.

The second response appeals to pluralism about causal structure. In fact, any causal theory of representation needs to be pluralistic about the types of causal correlation that facilitates representation. \cite{Stampe1977-STATAC}  gives an example of barometers: the barometric representation is caused by a drop in air pressure, and the drop in air pressure causes the storm, and in this way the barometer represents the storm. While the barometric reading represents the storm, the storm is neither a cause of the barometric reading nor one of its effects. This is one of the ways in which informational theories of representation improve on naive causal theories: they can allow for a wide range of causal relations to produce genuine representation. 

Another version of the sensory grounding challenge is illustrated by `The Octopus Test', a thought experiment from \cite{bender-koller-2020-climbing}. Imagine two humans communicating with one another through telegraphic cables across two islands, and a hyper-intelligent octopus that eavesdrops on their communication. In the beginning, the two humans mostly make small talk and describe their environments, which the octopus cannot see. However, because the octopus is hyperintelligent, it can pick up on various patterns in their communication. It has never seen trees, but it knows the patterns of discourse around the word ``tree'' and can do a good job of mimicking the other person. However, imagine that one of the people invents a catapult, and talks about it to their partner. 

The skeptical challenge claims that the octopus cannot represent the catapult, or any other object that only exists on the islands rather than in the octopus's own environment. Because the catapult is novel to the discourse, the octopus won't be able to understand what it is or make predictions about how it works. And if this is the case, it suggests that LLMs cannot represent the world either. 

We ourselves don't have a strong judgment about what the octopus can represent in this thought experiment. But we think one fruitful way of resolving the question is to consider various candidate conditions on representation. First, it is clear that the octopus carries information about the catapult, since it can reliably predict who will say what about the catapult. But, second, it is unclear whether the octopus has internal representations of the catapult that causally influence its predictions. It is also unclear whether the octopus has a series of internal representations of the catapult whose internal structure or pattern of similarity mirrors patterns in the catapult. Moreover, it is unclear whether the octopus evolved to track the catapult through some kind of optimization process such as natural selection. For all of these reasons, we think it is quite unclear whether the octopus represents the catapult.

Finally, it is unclear whether the octopus can reason well about the catapult in a wide range of counterfactual scenarios, in the kinds of ways that would produce genuine world models. The claim that the octopus simply could not in principle represent the catapult in such a way ignores a crucial hypothesis. When an LLM (or octopus) receives text, if it is sufficiently capable, it can form hypotheses over what sort of processes generated such a text. 

Just as scientists form hypotheses about microscopic phenomena responsible for observed patterns, advanced octopuses could make educated guesses about how the world works based on the patterns and types of text they receive. For example, if it reads lots of texts about machine learning, it might form certain hypotheses about certain types of researchers and computational infrastructure existing in the real world. From the under-water version of Wikipedia, it can make guesses about the sorts of processes and agents that would generate Wikipedia articles and what sort of underlying reality would result in such strings of text. 

While direct empirical evidence for LLMs (and octopuses) forming explicit hypotheses about the world is limited, the possibility remains theoretically plausible. If advanced LLMs can form hypotheses about how the world works, then the symbols they receive could connect to the outside world, even if they do not have the capability of receiving anything other than text as input. Future research should aim to uncover more about the internal mechanisms of LLMs and their ability to form and use such hypotheses about the world.

Another problem for the sensory grounding challenge is its fragility. Even if skeptics can successfully argue that a pure LLM is unable to represent reality, their victory may only be a Pyrrhic one. The transformer architecture, which underlies these models, is designed for sequence prediction generally and is versatile enough to be applied to various modalities, including computer vision.

Vision-Language Models (VLMs), such as GPT-4V, integrate both visual and textual data, processing them within a shared embedding space. Unlike systems that separately handle text and images and then attempt to combine their outputs, VLMs use a single, unified transformer model capable of processing and understanding both modalities simultaneously.

For instance, an image of a cat and the word ``cat'' would be represented in the same space, allowing the model to draw connections between the visual and textual representations. This shared embedding space helps bridge the gap between sensory input and linguistic representation. By incorporating visual data, VLMs gain a form of sensory grounding that pure text-based LLMs lack.

The architecture of VLMs is fundamentally similar to that of standard LLMs, relying on the same principles of self-attention and sequence modeling. This similarity undermines the argument that the transformer architecture itself is incapable of genuine representation. If VLMs, which are built on the same foundational architecture, can achieve sensory grounding and representational capabilities, it suggests that the perceived limitations of LLMs are not inherent to the architecture but rather a consequence of the input modalities used.

For all of these reasons, we do not consider the sensory grounding challenge a decisive threat to LLM representation. 

\subsection{Stochastic Parrots}

Another skeptical challenge to LLM representation is associated with the ``stochastic parrots'' argument, originally proposed by \cite{bender2021dangers}. This view contends that LLMs, despite their impressive outputs, are merely sophisticated statistical pattern matchers rather than systems capable of genuine understanding or representation.
The core claim of the stochastic parrots argument is that LLMs are trained solely to predict the next word in a sequence, without any true comprehension of the content they generate. According to this view, LLMs don't represent or reason about the world; instead, they simply reproduce patterns from their training data in a statistically sophisticated but fundamentally meaningless way. 
Proponents of this view often point to cases where LLMs produce fluent but nonsensical or contradictory outputs. For instance, an LLM might confidently assert a false statement or agree with contradictory premises in different conversations. These behaviors, they argue, reveal that LLMs lack genuine understanding and are merely ``parroting'' patterns from their training data.

One way to understand the stochastic parrot challenge is in terms of the teleosemantic conditions on representation we discussed earlier. Those conditions required that the internal states of LLMs have the function of representing features of the external world. This function would itself need to emerge from some kind of evolutionary, selective process. 

Here, we'll lay out four challenges for the stochastic parrots view:
The first challenge is overgeneralization. The argument that systems trained on pattern recognition can't develop genuine understanding potentially overgeneralizes. Human cognition, for instance, involves significant pattern recognition and statistical learning, yet we don't deny humans the capacity for genuine understanding. Similarly, other AI systems trained on pattern recognition (like computer vision models) are often considered to represent features of the world. The key question is not whether a system is trained on patterns, but whether it develops more sophisticated capabilities as a result of this training.

The second challenge is representation as a means to an end. While LLMs are indeed trained to predict the next word, representing the world could be an efficient means to this end. For example, to accurately predict words in a physics textbook, it would be helpful for an LLM to develop some internal understanding of physics concepts. Our earlier discussions of probing studies and world models provide evidence that LLMs do develop structured internal representations that go beyond simple pattern matching. (See \cite{herrmann2024standards} for more.) Again, this can be unpacked in terms of teleosemantic requirements on representation. The idea would be that the training process of predicting the next word selects for the ability to track features of the external world. 

The third challenge is emergent capabilities. Recent research suggests that LLMs can exhibit capabilities that were not explicitly part of their training objective. For instance: a) Few-shot learning: \cite{brown2020language} demonstrated that GPT-3 can perform new tasks with just a few examples, despite not being explicitly trained for this ability. This ``few-shot'' learning suggests a form of rapid adaptation that isn't easily explained by simple pattern matching. b) In-context learning: \cite{wei2022chain}  showed that LLMs can learn to perform new tasks from instructions and examples provided in the input prompt, without any change to their weights. This ability to ``learn'' within the context of a single forward pass challenges the notion of mere pattern reproduction. c) Chain-of-thought reasoning: \cite{wei2022chain} also demonstrated that prompting LLMs to generate step-by-step reasoning significantly improved their performance on complex reasoning tasks, suggesting a capacity for structured thinking beyond simple pattern matching. d) Emergence of coding abilities: \cite{chen2021evaluating}  found that large language models trained on natural language can develop unexpected coding abilities, despite not being explicitly trained on programming tasks. e) Zero-shot task generalization: \cite{kojima2022large} showed that LLMs can solve novel tasks they weren't explicitly trained on when prompted to ``think step by step,'' demonstrating a form of reasoning capability.

The fourth challenge is internal representations. As we've discussed earlier in this paper, interpretability research provides evidence that LLMs develop rich internal representations of concepts and relationships. This structured internal knowledge is difficult to reconcile with the stochastic parrots view.
It's worth noting that the stochastic parrots argument raises important questions about the nature of understanding and representation. Even if we reject the strong claim that LLMs are merely stochastic parrots, we might consider intermediate positions. For instance, LLMs might combine aspects of statistical pattern matching with more sophisticated representational capabilities.
Ultimately, the stochastic parrots argument highlights the need for careful empirical investigation of LLM capabilities and limitations. While the behavior of LLMs can sometimes be consistent with sophisticated pattern matching, the evidence we've reviewed throughout this paper suggests that LLMs do develop meaningful internal representations and capabilities that go beyond simple parroting.

\subsection{Memorization}

Another skeptical challenge concerns memorization. According to this challenge, we don't need to posit genuine reasoning in LLMs, because we can explain their behavior in another way. Instead of reasoning about questions, LLMs simply memorize great quantities of data and then use shallow heuristics to generalize to new prompts. They are trained on billions of sentences, and in the course of this training, they simply store the answers to a large number of questions. These answers are then returned in response to prompts, without any genuine reasoning taking place. 

Whether memorization threatens LLM representation depends on what is required for representation. If representation is simply a matter of carrying information, then memorization is perfectly compatible with genuine LLM representation. On the other hand, if representation requires the presence of robust models or of reasoning about the world that matches folk psychology, then memorization may rule out representation.

In fact, there is a lot of evidence that LLMs do not merely memorize the answers to questions they are asked. Instead, LLMs seem to possess the ability to generalize robustly from their training data and then make correct predictions about new questions. They do not simply use shallow heuristics. 

Return to the case of modular addition. As we discussed above, \cite{nanda2023progress} found that during training, the LLM learns modular addition in multiple steps: an initial period of overfitting, based on memorization, followed by a transition to a general solution to the problem. In the initial phase of training, the LLM does engage in memorization. But after using memorization, the LLM learns a more general algorithm for computing the answer. After this algorithm is learned, the LLM then removes its memorization components. As evidence for this claim, Nanda et al found that in the initial period of training, the model achieved 100\% accuracy on the training data, but low accuracy on the testing data. After 10,000 epochs of training, the model learned how to actually perform the task, and achieved high accuracy on the testing data. Overall, this suggests that LLM skeptics are missing out on much of the rich structure of LLM reasoning. 

Similar abilities to generalize were found in the Othello experiment. Every game of Othello starts with one of four moves. Each initial move creates a `quadrant' of Othello game space. To train Othello, the experiments only included training data from 3 of the 4 quadrants. But they found that the model was equally successful at playing Othello in the omitted quadrant of game space. \footnote{Our focus in this paper is on whether LLMs have a folk psychology. This question is worth distinguishing from another skeptical target: the question of whether LLM \emph{outputs} are meaningful. Here, the key question is whether text produced by LLMs has meaning. This is a different question than whether LLMs have a folk psychology. Compare: we can imagine a human being who has beliefs and desires, but who does not know how to speak French. If they started saying French sentences out loud phonetically, there would be an interesting question to ask about whether these sentences are meaningful in their mouth. But this question is a separate one from whether the speaker has a psychology.  In practice, much of the debate about LLM outputs has itself been connected to debates about LLM folk psychology. For example, one skeptical challenge to the meaningfulness of LLM outputs starts from the premise that LLMs lack communicative intentions. This is itself a skeptical premise about LLM folk psychology. In response, \cite{mandelkern2023language} have suggested that LLM psychology may not be necessary for LLM outputs to be meaningful, as long as LLMs as part of our linguistic community. (See also \cite{mollo2023vectorgroundingproblem}.) Others have suggested that we think of LLMs as more like \emph{libraries} rather than speakers, which could also allow outputs to be meaningful (for discussion, see \cite{gopnik2022children}). By contrast, \cite{lederman2024language} argue that some kinds of LLM outputs can only be meaningful of LLMs are genuine speakers. Our focus in this paper is the question of whether LLMs have a psychology, rather than the question of whether LLM outputs are meaningful.} 
\footnote{%
    The connection between belief and learning may offer another route to LLM skepticism. The philosopher Grace Helton has argued that in order to genuinely have beliefs, those beliefs must be able to change in response to evidence (\cite{Helton2018-HELIYC}). Helton's argument has two premises. First, you have a belief only if you are obligated to change that belief in response to strong counter-evidence. Second, you are obligated to do something only if you are able to do so. From these premises, it follows that LLMs have beliefs only if they are able to change their beliefs in response to strong counter-evidence. But one might worry that after training, LLMs can no longer learn or change their beliefs. If Helton's premises are correct, it follows that LLMs do not have beliefs.

    The initial skeptical concern here is that after training LLMs, no longer learn or change their beliefs. The idea is that all of the learning that an LLM engages in occurs during training, when the weights of the LLM's neural network gradually change in response to feedback. But once a user interacts with the LLM, for example using ChatGPT, the weights are frozen, and so the LLM no longer learns.

    This skeptical response fails, because of in-context learning. In-context learning refers to the ability of a language model to use the context provided within a given prompt or conversation to generate relevant and coherent responses (\cite{brown2020language}). In general, in-context learning can refer to any way in which the model aptly adapts to the context of the prompt: e.g., adapting the right style, responding to corrections, or maintaining continuity over a conversation.

    For our purposes, we can focus on few-shot learning, which is a form of in-context learning. In few-shot learning, you can provide examples within your prompt, and the model will use these to understand the type of response you're looking for. 

    For example, if you want GPT-3 to predict nationalities, you might just ask it about Marie Curie's nationality. But with a few-shot prompt, you instead give it Einstein and Gandhi's nationality, and then ask it Marie Curie's nationality. Often, LLMs will perform better after seeing a few examples, rather than answering a question `zero shot'. This suggests that the LLM learns from the initial examples (\cite{xie2021explanation}). 

    \cite{xie2021explanation} found that in-context learning in LLM satisfies many features of Bayesian inference. In particular, they hypothesized that in few-shot learning, there's a hidden concept that explains each of the examples in the prompt and that should be used to generate the response. They studied how a perfect bayesian would guess the right response given this hypothesis and compared it to how GPT-2 performed. GPT-2 learned quickly, like a bayesian, and got better and better with more examples. 

    Although LLMs of today do not retain the information learned in-context in other inference cycles, they do update their beliefs within a given inference cycle. Humans likewise often forget some of their beliefs–albeit not always–so we do not see a special reason to deny that LLMs can learn and change their minds. 

    Philosophers often distinguish dispositional from occurrent beliefs. This distinction is particularly interesting in the setting of LLMs. In LLMs, dispositional representations would be stored in the underlying weights of the neural net, while occurrent representations would be tokened by the activations in response to prompts. When LLMs engage in on-line learning, their occurrent representations change in response to evidence. In this way, Helton's argument could allow that LLMs possess occurrent beliefs, even if they lack dispositional beliefs. Rather, they would possess unchanging dispositional representations that do not respond to evidence, and so are not subject to rational obligations, even though they could guide action and inference.
    } %

\section{Action and Folk Psychology}
\label{sec:folk_psych}

While we've argued that LLMs have mental representations, the question remains: do they have a robust folk psychology? To address this, we need to consider whether LLMs have beliefs about the world, desires they aim to satisfy, and intentions that guide their actions. This question is complex, involving issues of stability, coherence, and goal-directed behavior.

To approach this question, we'll examine two influential philosophical perspectives on folk psychology: interpretationism and representationalism. Each offers a different framework for understanding mental states and presents different challenges when applied to LLMs.

\subsection{Interpretationism}

The most radical view is interpretationism. The idea behind interpretationism is that folk psychology is for explaining behavior.  All that is required to have a folk psychology is for the system to behave in sufficiently complex ways best explained by appeal to folk psychological states. When this happens, the system has beliefs and desires: the system's  desires are the goals promoted by its actions, and its beliefs are the views about the world that are required to be true in order for its actions to promote its goals. Whether it has internal states of a certain kind is not directly relevant to the question of folk psychology. Prominent interpretationists include \cite{Dennett1981-DENTIS} and \cite{Davidson1984-DAVIIT}.\footnote{%
    Closely related to interpretationism is dispositionalism. On a dispositionalist picture, an agent has a belief that P if it is disposed to behave as if P. See, e.g., \cite{Marcus1990-MARSRP-2}.
} %

The previous conditions we've explored could allow for a separation between mental representation in general and belief/desire psychology in particular. LLMs could satisfy informational, causal, structural, and teleosemantic requirements on content, even if their behavior is too disorganized and happenstance to count as possessing beliefs and desires. Importantly, for interpretationists, folk psychological states come as a package deal. As \cite{Stalnaker1984-STAI} puts it:

\begin{quote}Belief and desire \ldots are correlative dispositional states of a potentially rational agent. To desire that P is to be disposed to act in ways that would tend to bring it about that P in a world in which one's beliefs, whatever they are, were true. To believe that P is to be disposed to act in ways that would tend to satisfy one's desires, whatever they are, in a world in which P (together with one's other beliefs) were true. (p. 15)\end{quote}

So, even if LLMs have something like mental representations of the world, they might not have desires or the right kinds of behavioral dispositions required for folk psychological states.

Do large language models satisfy interpretationist conditions on mental representation? There are at least two reasons for skepticism. The first challenge concerns LLM affordances: what actions an LLM can perform. Pure LLMs do not have access to robotic bodies. Instead, they simply produce text (or probability distributions over tokens). But at first glance these text interactions aren't naturally suited for performing complex actions.

There are at least two good responses to the problem of affordances.  First, the problem of affordances is fragile. Some LLMs today do have access to robotic limbs. For example, Google's Palm-E system integrates an LLM with a robotic limb, which can perform actions after being prompted by a user with text (\cite{driess2023palme}). When asked to pick up a bag of potato chips from the counter, the robotic limb can skillfully sort between different objects and pick up the chips.

Second, text production provides a rich enough space of alternative outcomes to count as a full-fledged action space. In today's digital world much human action takes place in text. When we imagine a human being whose life is confined entirely to text-based actions, we see no barrier to such a being possessing beliefs and desires. To see this point in greater detail, consider the use of LLMs in game environments. \cite{mei2024turing} studied the behavior of LLMs in social cooperation games. These kinds of environments allow LLMs to formulate complex plans. They found that GPT-3 and GPT-4 would adjust their behavior throughout the game: 
\begin{quote}
    In games with multiple roles (such as the Ultimatum Game and the Trust Game), the AIs' decisions can be influenced by previous exposure to another role. For instance, if ChatGPT-3 has previously acted as the responder in the Ultimatum Game, it tends to propose a higher offer when it later plays as the proposer, while ChatGPT-4's proposal remains unchanged. Conversely, when ChatGPT-4 has previously been the proposer, it tends to request a smaller split as the responder. (p. 17)
\end{quote} 
This kind of behavior suggests that LLMs have the ability to use game theory to navigate environments involving other agents. 

A second reason for skepticism about action is instability. Shanahan and others have noted that LLM outputs are very sensitive to prompting. If you slightly change the way you ask a question, the LLM can start to behave very differently. This makes it hard to see the LLM as taking a wide range of means to promote a unified goal (\cite{shanahan2023role}). 

We think instability is potentially the most serious threat to LLM folk psychology. We see two potential responses worthy of further development. First, one might concede that LLMs are relatively unstable but say that each prompting session with an LLM produces a different agent with its own beliefs and desires. For example, if you ask GPT-4 to write poetry for you on one day and start a new conversation asking it to play chess with you the second day, you in effect have two separate agents. The chess-playing instance has no interest in poetry, but it does have the goal of beating you at chess. When you give it goals like writing poetry or playing chess or coding, it does a good job. Those tasks are very difficult. So, the thought goes, the best explanation is that the beliefs and desires of the LLM change from inference cycle to inference cycle, but the behavior of particular instances is still fruitfully explained with beliefs and desires and intentions.

Second, one might reject the claim of instability. Although the behavior looks unstable, the different behavior of LLMs in different prompting sessions might derive from a stable underlying goal, such as pleasing the user. The different outputs of LLMs in different sessions might result from a disconnect between what the model believes and what the model says. In other words, instability in model outputs is evidence of lying, not evidence that the system lacks beliefs.\footnote{
    For an argument that RLHF helps ground textual meaning through the goal of pleasing the user, see \citep{mollo2023vectorgroundingproblem}.
} 

Above all, we suggest that more research needs to be done about how stable LLM outputs are to a range of different prompting environments. For example, one fruitful project would be to explore how LLM strategies in game environments change as a result of different prompting conditions. This would allow some measure of whether it forms coherent plans that are robust to a range of perturbations.

\subsection{Representationalism}

Representationalism, advocated by philosophers like Jerry Fodor and Fred Dretske, holds that having mental states requires having internal representations with appropriate functional roles. On this view, to have a belief that p, a system must have an internal state that represents p and plays the right causal role in the system's cognitive economy.

There is cause for optimism on the representationalist picture. As we've argued at length, some LLM mental states can have truth-conditions. Furthermore, LLMs even appear to have some sorts of world models and structured representations of conceptual domains.\footnote{%
    We might require that LLMs internally distinguish between true claims and false claims in a way that goes beyond representation for the LLM to count as having beliefs. In particular, we might require that they somehow systematically ``tag'' sentences as true or false and use this tag in their master algorithm to determine what text to output? 

    We don't yet have a full answer to this question. Some (\cite{burns2024discovering}, \cite{azaria2023internal}) have argued that LLMs do make an internal distinction between truth and falsity. However, others (e.g., \cite{levinstein2024still}) claim these studies are flawed. \cite{herrmann2024standards} argue for a representationalist account of belief for LLMs but maintain that current empirical evidence of whether LLMs actually have beliefs is inconclusive.} 

However, even on representationalist pictures like Fodor's, beliefs have to play the right role in the larger system and generally cannot be divorced entirely from desires. Even if LLMs have rich internal representations, it's not clear that these play the right kind of role in generating behavior to count as beliefs or desires. The issue of instability resurfaces here---if internal representations don't stably guide behavior across different prompts, can they really count as beliefs?

Ultimately, we think matters are easier for the representationalist than the interpretationist. While we have some behavioral evidence in the case of LLMs, behavioral evidence is much more limited for LLMs than it is for humans. However, we have perfect internal access to LLMs, and we have much to discover about the role various representations play in LLMs' cognition. Therefore, as we come to understand more about how LLMs think, it will become more obvious for representationalists whether they have folk psychological mental states.
Some key open questions for attributing folk psychological states on either picture, then, include:

\begin{itemize}
\item \textbf{Action and planning}: How can we best understand LLM ``action'' given their limited affordances?
\item \textbf{Stability}: How can we reconcile the apparent instability of LLM outputs with the need for stable beliefs and desires?
\item \textbf{Goal-directedness}: Do LLMs have anything analogous to enduring goals or values?
\end{itemize}

Addressing these questions will require a combination of philosophical analysis and empirical investigation. While there's evidence that LLMs have sophisticated internal representations and can exhibit complex, apparently goal-directed behavior, significant questions remain about whether they possess full-fledged folk psychological states. Resolving these questions will be crucial for understanding the capabilities, limitations, and potential moral status of these increasingly ubiquitous AI systems.

\bibliographystyle{plainnat}
\bibliography{llm}

\end{document}